\documentclass[conference]{IEEEtran}
\IEEEoverridecommandlockouts
\usepackage{cite}
\usepackage{amsmath,amssymb,amsfonts}
\usepackage{graphicx}
\usepackage{textcomp}
\usepackage{xcolor}
\usepackage{amsmath}
\usepackage{algorithmicx}
\usepackage{booktabs}
\usepackage{diagbox}
\usepackage{url}
\def\BibTeX{{\rm B\kern-.05em{\sc i\kern-.025em b}\kern-.08em
    T\kern-.1667em\lower.7ex\hbox{E}\kern-.125emX}}
\begin{document}

\title{CCasGNN: Collaborative Cascade Prediction Based on Graph Neural Networks
}

\author{\IEEEauthorblockN{1\textsuperscript{st} Yansong Wang}
\IEEEauthorblockA{\textit{College of Computer and Information Science} \\
\textit{Southwest University}\\
Chongqing, China \\
yansong0682@email.swu.edu.cn}
\and
\IEEEauthorblockN{2\textsuperscript{nd} Xiaomeng Wang}
\IEEEauthorblockA{\textit{College of Computer and Information Science} \\
\textit{Southwest University}\\
Chongqing, China \\
wxm1706@swu.edu.cn}
\and
\IEEEauthorblockN{3\textsuperscript{rd} Tao Jia}
\IEEEauthorblockA{\textit{College of Computer and Information Science} \\
\textit{Southwest University}\\
Chongqing, China \\
tjia@swu.edu.cn, corresponding author}
}

\maketitle

\begin{abstract}
Cascade prediction aims at modeling information diffusion in the network. Most previous methods concentrate on mining either structural or sequential features from the network and the propagation path. Recent efforts devoted to combining network structure and sequence features by graph neural networks and recurrent neural networks. Nevertheless, the limitation of spectral or spatial methods restricts the improvement of prediction performance. Moreover, recurrent neural networks are time-consuming and computation-expensive, which causes the inefficiency of prediction. Here, we propose a novel method CCasGNN considering the individual profile, structural features, and sequence information. The method benefits from using a collaborative framework of GAT and GCN and stacking positional encoding into the layers of graph neural networks, which is different from all existing ones and demonstrates good performance. The experiments conducted on two real-world datasets confirm that our method significantly improves the prediction accuracy compared to state-of-the-art approaches. What’s more, the ablation study investigates the contribution of each component in our method.
\end{abstract}

\begin{IEEEkeywords}
information cascade, cascade prediction, collaborative prediction, graph neural networks
\end{IEEEkeywords}

\section{Introduction}
Social networks such as Twitter, Weibo, and YouTube have greatly facilitated our life, in which people post what they see and hear with friends \cite{keim2011emergent}. Government institutions are using platforms to communicate with their communities \cite{mickoleit2014social}. Thus, understanding the information dissemination mechanism is valuable for us to discover hot information in advance.

A plethora of methods have been proposed to solve the problems of cascade prediction. Traditional methods relying on hand-crafted features inherit a high interpretability of feature engineering \cite{tsur2012s,cheng2014can}. However, the feature based models suffer from the issue of scalability, which obstruct their implementation. In the cascade, the spreading of information can be formulated as node sequence in the continuous temporal domain. Therefore, methods that model the arrival of event sequence are introduced to characterize the key factors in the information diffusion process \cite{shen2014modeling, bao2015modeling}. Despite an enhanced prediction accuracy in some cases, such methods are unable to fully leverage the implicit information in the cascade dynamics and require long observation dependency. With the successful application of deep learning and representation learning in various fields \cite{zhan2021coarsas2hvec, perozzi2014deepwalk}, many recent approaches start to consider neural networks and achieve good performance. In deep learning based approaches, DeepWalk \cite{perozzi2014deepwalk} and graph neural networks (GNNs) are generally applied to sample the topological information, while the recurrent neural network (RNN) or its variants such as Bi-GRU \cite{chung2014empirical}, LSTM \cite{hochreiter1997long} are introduced to capture the temporal features. However, both spectral and spatial GNNs have their own limitations. For example, spectral methods are not suitable for inductive tasks and spatial methods lack sufficient theoretical derivation. Meanwhile, recurrent neural networks are time-consuming and computation-expensive, which is not suitable for time-sensitive tasks. Hence, how to overcome such limitations becomes an interesting and ongoing challenge.

In this paper, we propose a collaborative cascade prediction framework based on graph neural networks (CCasGNN) to jointly utilize the structural features, sequence information, and user profiles. Specifically, CCasGNN models user embedding through collaborative work of the graph attention network (GAT) \cite{velivckovic2017graph} and the graph convolutional network (GCN) \cite{kipf2016semi}. We also stack positional encodings into the layers to ensure that positional information should be considered in every graph neural network layer. The multi-head attention mechanism is employed to capture the relationships between all users from a global perspective. We then aggregate embeddings by average pooling and obtain the results through multilayer perceptron and weighted sum. The code is publicly available at \url{https://github.com/MrYansong/CCasGNN} for future reference and reproducibility.

\section{Related Work}
At present, the research on the prediction of information popularity is mainly divided into two aspects: Classification \cite{gou2018learning, liao2019popularity} and Regression \cite{khosla2014makes,li2017deepcas,cao2017deephawkes,chen2019information}. We now review the related literature as follows.

\subsection{Feature based approaches}
Models based feature engineering select feature sets that have a great impact on propagation according to the analysis of historical dissemination. \cite{tsur2012s} proposes an efficient linear regression model based on the content features for predicting the spread of an idea. \cite{cheng2014can} implements machine learning methods on a large sample of features and finds that temporal and structural features are key predictors of cascade size. \cite{elsharkawy2016towards} points that feature governing the information popularity vary from one dataset to another and the excellent preselection of features will improve the accuracy of the prediction task. Therefore, generalization ability is an essential factor that we must consider based on these approaches.

\subsection{Process based approaches}
Information retweeting can be modeled as the occurrence sequence generated by the underlying dynamics of information diffusion. Therefore, various stochastic process based models are used to characterize the spreading of information. The reinforced Poisson process was used in \cite{shen2014modeling} to predict the popularity of individual items, in which three key ingredients (fitness, temporal decay, and reinforcement mechanism) are introduced. \cite{bao2015modeling} presents a popularity prediction model based on the self-excited Hawkes process, which distinguishes the incentive size of each forwarding and improves the performance to a certain extent. The review \cite{gao2019taxonomy} indicates that Poisson process is too simple to capture the propagation patterns, and Hawkes process usually overestimate the popularity, probably due to its rudimentary self-excitation mechanism. Although these methods are mathematically rigid, the strong assumptions on the underlying process limit their generality and accuracy.

\subsection{Deep learning based approaches}
Inspired by the successful application of deep learning in various fields, researchers attempt to improve the performance of cascade prediction by neural networks. Representation learning and deep learning are used in \cite{khosla2014makes} to predict the number of views an image will receive. DeepCas \cite{li2017deepcas} borrows the idea of random walks to sample the node sequence and learns node embedding by DeepWalk \cite{perozzi2014deepwalk}. The sequences are fed into bidirectional gated recurrent units (Bi-GRU) \cite{chung2014empirical} along with attention mechanisms \cite{bahdanau2014neural} to obtain the information popularity. DeepHawkes \cite{cao2017deephawkes} attempts to learn node representation by a supervised framework with the interpretable Hawkes process and generates node sequence based on the propagation cascade for predicting. CasCN \cite{chen2019information} samples a series of sequential subcascades and adopts a dynamic multi-directional GCN \cite{kipf2016semi} to learn structural information. In addition, LSTM \cite{hochreiter1997long} is used to capture the temporal features. Although satisfactory performance we obtained, how to overcome the limitations of GNNs and RNN is still worth researching.

\section{Preliminaries}
\textbf{Cascade Graph}:
Let $G=\left( V,E \right)$ be a static social network, where $V$ denotes the set of users and $E\subseteq V\times V$ denotes the set of edges. Suppose we have $n$ messages, each piece of information ${{m}_{i}}$ spreading in the network will form the information cascade ${{C}_{i}}=({{V}_{i}},{{E}_{i}},{{P}_{i}})$, where ${{V}_{i}}\subseteq V$ is a set of users that have been involved in the cascade ${{C}_{i}}$, ${{E}_{i}}\subseteq {{V}_{i}}\times {{V}_{i}}$ is a set of edges and ${{P}_{i}}$ is the diffusion path of vertices.

\textbf{Diffusion Path}:
In the process of information propagation, the diffusion path $P$ between users are created. The path of cascade ${{C}_{i}}$ can be represented as ${{P}_{i}}=\left\{ \left( {{v}_{1}},{{p}_{{{v}_{1}}}} \right),\ldots ,\left( {{v}_{j}},{{p}_{{{v}_{j}}}} \right) \right\}$, where ${{p}_{{{v}_{j}}}}$ denotes the position of vertex ${{v}_{j}}$ in the cascade ${{C}_{i}}$.

\textbf{Problem Definition}:
In our study, information popularity is defined as the number of incremental retweets $\Delta R_{i}^{T}$ of a message ${{m}_{i}}$ after the observation time window $\left[ 0,\text{T} \right]$. The $\Delta R_{i}^{T}$ can be denoted as $\Delta R_{i}^{T}=\left| R_{i}^{t+\Delta t} \right|-\left| R_{i}^{t} \right|$, when $\Delta t$ is large enough, $\left| R_{i}^{t+\Delta t} \right|$ can represent the final cascade size.

\begin{figure*}
	\includegraphics[width=\textwidth]{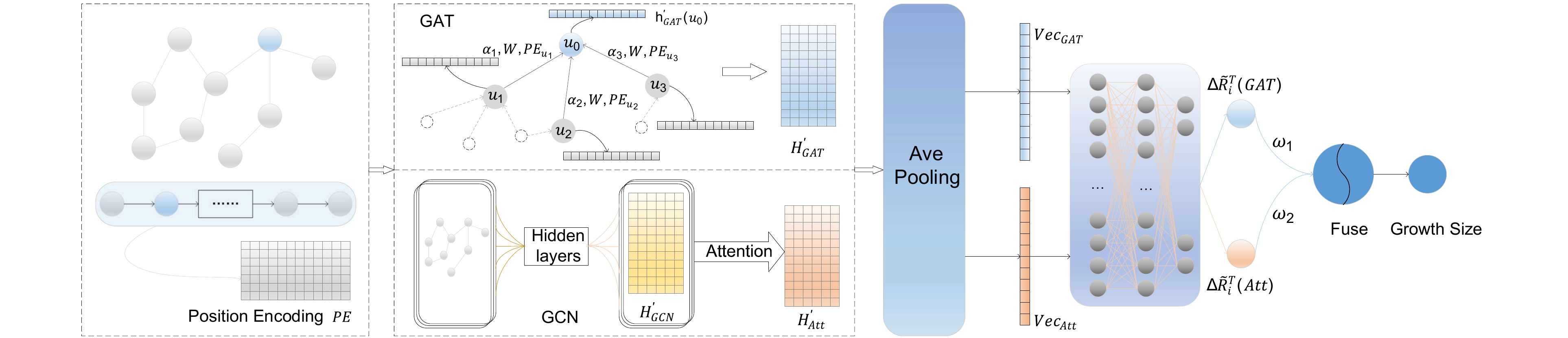}
	\caption{Framework of CCasGNN.}
	\label{fig:teaser}
\end{figure*}
\section{Method}
In this section, we introduce the components of our model in detail. The framework of CCasGNN is depicted in Fig. 1.

\subsection{Position Encoding}
To make use of the sequence order information without RNN, we introduce position encoding, aiming at encodings user position information in the embeddings. These encodings characterize the relative or absolute position in the propagation sequence and have the same dimension ${{d}_{p}}$. Following the work \cite{vaswani2017attention}, the positional encoding $PE\in {{\mathbb{R}}^{n\times {{d}_{p}}}}$ is calculated as:
\begin{equation}
P{{E}_{\left( {{p}_{{{v}_{j}}}},k \right)}}=\left\{ \begin{matrix}
	\sin \left( {\frac{1}{{{1000}^{2i/{{d}_{p}}}}}}\cdot {{p}_{{{v}_{j}}}} \right),~~~if~k=2i  \\
	\cos \left( {\frac{1}{{{1000}^{2i/{{d}_{p}}}}}}\cdot {{p}_{{{v}_{j}}}} \right),~~~if~k=2i+1  \\
\end{matrix} \right.,
\end{equation}
where ${{p}_{{{v}_{j}}}}\in R$ is the position of user ${{v}_{j}}$, and $k$ is the dimension of the positional encoding, ${{d}_{p}}$ is the total dimension of the positional encoding.

For any fixed offset $t$, $P{{E}_{{{p}_{{{v}_{j}}}}+t}}$ can be represented as a linear transformation of $P{{E}_{{{p}_{{{v}_{j}}}}}}$. Thus this method captures the relative position between users in the diffusion sequence with high space and computation efficiency.

\subsection{User Embedding}
In order to embed the structural, positional, and individual information into latent space, we utilize the graph neural network in our model. Specifically, GCN and GAT are implemented collaboratively to overcome the inherent limitations of spectral and spatial approaches. Since our model does not contain recurrent networks, the relative position of users in the diffusion sequence is stacked into the graph neural network layers. For a cascade ${{C}_{i}}$, the input to the GCN and GAT layers consists of two parts: a vertex feature matrix $H\in {{\mathbb{R}}^{n\times F}}$ and an adjacency matrix $A\in {{\mathbb{R}}^{n\times n}}$ of the cascade graph, where $n$ is the number of vertices, $F$ is the number of features. Each row of the feature matrix denotes the static profile of a user.

\subsubsection{GAT layer}
\ 
\par
GAT incorporates the attention mechanism into the embedding step, which computes the hidden states of each node by attending to its neighbors. This method can be seen as an aggregator, the hidden state of node $v$ at the ${{l}^{th}}$ layer can be obtained by:
\begin{equation}
	h_{GAT\left( v \right)}^{'}\left( l \right)=\text{ }\!\!\sigma\!\!\text{ }\left( \underset{u\in {{N}_{v}}}{\mathop \sum }\,{{\alpha }_{uv}}W(h_{GAT\left( u \right)}^{'}\left( l-1 \right)||P{{E}_{u}}) \right),
\end{equation}
where $P{{E}_{u}}\in {{\mathbb{R}}^{{{d}_{p}}}}$ is the positional encoding of user $u$, $h_{GAT\left( u \right)}^{'}$ is equal to ${{h}_{u}}\in {{\mathbb{R}}^{F}}$ in the first layer, $W$ is a shared weight transformation vector, $N_v$ denotes the neighbors of user $v$, the coefficient ${{\alpha }_{uv}}$ implicates the contribution of user $u$ to user $v$,
\begin{equation}
	{{\alpha }_{uv}}=\frac{\text{exp}(LeakReLU({{a}^{\intercal }}[W{{h}_{v}}||W{{h}_{u}}]))}{\mathop{\sum }_{k\in {{N}_{v}}}\text{exp}(LeakyReLU({{a}^{\intercal }}[W{{h}_{v}}||W{{h}_{k}}]))},
\end{equation}
where $W$ is the weight matrix associated with the linear transformation to each node, and $a$ is the weight vector.

\subsubsection{GCN layer}
\ 
\par
GCN studies the properties of graphs with the help of eigenvalues and eigenvectors of graph Laplace matrices. We denote the hidden representation output of GCN at the ${{l}^{th}}$ layer as follows:
\begin{equation}
H_{GCN}^{'}\left( l \right)=\sigma ({{L}^{sn}}(H_{GCN}^{'}\left( l-1 \right)|\text{ }\!\!|\!\!\text{ }PE\text{)}{{W}^{\intercal }}+b),
\end{equation}
where $PE\in {{\mathbb{R}}^{n\times {{d}_{p}}}}$ is the positional encodings, $W$ is the learnable weight parameters, $\text{ }\!\!\sigma\!\!\text{ }$ is the activation function, $H_{GCN}^{'}$ is equal to $H\in {{\mathbb{R}}^{n\times F}}$ in the first layer, ${{L}^{sn}}$ is the symmetric normalized Laplace matrix defined as:
\begin{equation}
{{L}^{sn}}={{D}^{-1/2}}L{{D}^{-1/2}}=I-{{D}^{-1/2}}A{{D}^{-1/2}},
\end{equation}
where $L$ is the Laplace matrix, $D$ is the degree matrix (diagonal matrix), and $I$ is the identity matrix.

\subsubsection{Attention mechanism}
\ 
\par
To further capture the interactions between users, we perform the multi-head attention \cite{vaswani2017attention} after GCN layers. Instead of implementing only one scaled dot-product attention, there are multiple versions of queries, keys, and values performing the attention function in parallel, and yielding ${{d}_{v}}$-dimensional output values. Here, we take one-head attention mechanism as an example. Firstly, query matrix $Q\in {{\mathbb{R}}^{n\times {{d}_{q}}}}$, key matrix $K\in {{\mathbb{R}}^{n\times {{d}_{k}}}}$, and value matrix $V\in {{\mathbb{R}}^{n\times {{d}_{v}}}}$ will be created,
\begin{equation}
\begin{split}
\text{Q}=H_{GCN}^{'}{{W}^{Q}},~~
\text{K}=H_{GCN}^{'}{{W}^{K}},~~
\text{V}=H_{GCN}^{'}{{W}^{V}},
\end{split}
\end{equation}
where $H_{GCN}^{'}\in {{\mathbb{R}}^{n\times {F}'}}$ is the output of GCN layers, ${{W}^{Q}}\in {{\mathbb{R}}^{{F}'\times {{d}_{q}}}}$, ${{W}^{K}}\in {{\mathbb{R}}^{{F}'\times {{d}_{k}}}}$, and ${{W}^{V}}\in {{\mathbb{R}}^{{F}'\times {{d}_{v}}}}$ are projection parameter matrices. Then the attention coefficient matrix $Att\in {{\mathbb{R}}^{n\times n}}$ is defined as follows:
\begin{equation}
Att=softmax\left( \frac{Q{{K}^{\intercal }}}{\sqrt{{{d}_{k}}}} \right),
\end{equation}
the purpose of dividing by $\sqrt{{{d}_{k}}}$ is to counteract vanishing gradient problems. Finally, we formula the matrix of outputs as follows:
\begin{equation}
H_{Att\_one}^{'}\left( Q,K,V \right)=softmax\left( \frac{Q{{K}^{\intercal }}}{\sqrt{{{d}_{k}}}} \right)V,
\end{equation}
where $H_{Att\_one}^{'}\in {{\mathbb{R}}^{n\times {{d}_{v}}}}$.

Multi-head attention leverages the parallelism of attention to divide the model into multiple subspaces so that it can focus on the information at different positions. It is defined as:
\begin{equation}
\begin{split}
H_{Att}^{'}\left( Q,K,V \right)=Aggregate\left( hea{{d}_{1}},\ldots ,hea{{d}_{h}} \right),\\
hea{{d}_{i}}=H_{At{{t}_{i}}}^{'}\left( QW_{i}^{Q},KW_{i}^{K},VW_{i}^{V} \right),
\end{split}
\end{equation}
where $W_{i}^{Q}\in {{\mathbb{R}}^{{{d}_{att}}\times {{d}_{q}}}}$, $W_{i}^{K}\in {{\mathbb{R}}^{{{d}_{att}}\times {{d}_{k}}}}$, and $W_{i}^{V}\in {{\mathbb{R}}^{{{d}_{att}}\times {{d}_{v}}}}$ are parameter matrices. There are many choices for pooling, we select average operation in this paper.

\subsection{Prediction}
The $H_{GAT}^{'}$ and $H_{Att}^{'}$ obtained from the above two parts are further aggregated as vectors that can be fed into multi-layer perceptrons (MLP) to output two components of the prediction result.
\begin{equation}
\begin{split}
Vec_{GAT}=Aggregate\left( H_{GAT}^{'} \right),\\
Vec_{Att}=Aggregate\left( H_{Att}^{'} \right),\\
\Delta \tilde{R}_{i}^{T}\left( GAT \right)=MLP\left( Vec_{GAT} \right),\\
\Delta \tilde{R}_{i}^{T}\left( Att \right)=MLP\left( Vec_{Att} \right),\\
\end{split}
\end{equation}
where $Ve{{c}_{GAT}}\in {{\mathbb{R}}^{{{d}_{GAT}}}}$ and $Ve{{c}_{Att}}\in {{\mathbb{R}}^{{{d}_{Att}}}}$ are vectors that represent the cascade graph, $\Delta \tilde{R}_{i}^{T}$ is the prediction result of a message ${{m}_{i}}$ after the observation time window $\left[ 0,\text{T} \right]$. We choose the simple averaging operation as the aggregation function.

The prediction result $\Delta \tilde{R}_{i}^{T}\left( GAT \right)$ and $\Delta \tilde{R}_{i}^{T}\left( Att \right)$ are combined to get the final cascade prediction. We use the weighted sum to finish this process,
\begin{equation}\label{e1}
\Delta \tilde{R}_{i}^{T}={{w}_{1}}\Delta \tilde{R}_{i}^{T}\left( GAT \right)+{{w}_{2}}\Delta \tilde{R}_{i}^{T}\left( Att \right),
\end{equation}
where the weight ${{w}_{1}}$ and ${{w}_{2}}$ are dynamically adjusted during the training process by error back propagation.
The loss function to be minimized is defined as:
\begin{equation}
\begin{split}
L=\frac{1}{N}\underset{i=1}{\overset{N}{\mathop \sum }}\,{{w}_{1}}{{L}_{GAT}}+{{w}_{2}}{{L}_{Att}},\\
{{L}_{GAT}}={{\left( \log \Delta \tilde{R}_{i}^{T}\left( GAT \right)-\log \Delta R_{i}^{T} \right)}^{2}},\\
{{L}_{Att}}={{\left( \log \Delta \tilde{R}_{i}^{T}\left( Att \right)-\log \Delta R_{i}^{T} \right)}^{2}},
\end{split}
\end{equation}
where $N$ is the total number of cascades. Following [15,18,19], we use the log value of the growth size.

\section{Experiments}
We compare the performance of CCasGNN with several competitive models in two real-world datasets. In addition, we perform extended ablation studies to validate the effectiveness of the components we proposed. Hyper-parameter sensitivity is also discussed.

\subsection{Datasets}
We use two real-world datasets, the Sina Weibo \cite{zhang2013social} and the DBLP citation network \cite{tang2008arnetminer}. The statistics of the datasets as given in Table I.
\begin{table}[htbp]
	\centering
	\renewcommand\arraystretch{0.9}
	\caption{ statistics of the datasets.}
	\setlength{\tabcolsep}{1.1mm}{
	\begin{tabular}{c|ccc|ccc}
		\toprule
		\multicolumn{1}{c}{DataSet} & \multicolumn{3}{|c|}{Weibo} & \multicolumn{3}{c}{DBLP} \\
		\midrule
		${T}$     & \multicolumn{1}{c|}{1 hour} & \multicolumn{1}{c|}{2 hours} & \multicolumn{1}{c|}{3 hours} & \multicolumn{1}{c|}{3 years} & \multicolumn{1}{c|}{5 years} & \multicolumn{1}{c}{7 years} \\
		\midrule
		cascades & 29,169 & 29,116 & 29091 & 30,106 & 29,998 & 29,991 \\
		Avg. nodes & 34.790 & 36.527 & 38.471 & 21.774 & 25.678 & 28.205 \\
		Avg. edges & 28.652 & 29.347 & 31.813 & 31.374 & 42.152 & 48.793\\
		Avg. growth & 178.841 & 122.819 & 95.522 & 27.949 & 14.887 & 6.432 \\
		\bottomrule
	\end{tabular}%
	}
	\label{tab:addlabel}%
\end{table}%

\textbf{Sina Weibo} is the most popular Chinese microblogging platform, which contains the following network, masked data of users, and retweet trajectories. Following \cite{chen2019information}, we extract all retweets of each cascade within the next 24 hours and filter out cascades with fewer than 10 retweets. The length $T$ of the observation time window is set as 1 hour, 2 hours, and 3 hours.

Following the previous work \cite{li2017deepcas,cao2017deephawkes,chen2019information}, we utilize the citation network to demonstrate the generalization of our model. \textbf{DBLP} dataset contains the paper title, publish time, abstract, and citation relationship. In order to build the paper profile, we firstly use Word2Vec \cite{mikolov2013efficient} to embed the words of the abstract into vectors and then average these vectors that describe the same paper as the profile representation. We extract the citation relationship within the next 10 years and filter out cascade sequences with lengths fewer than 10 in our experiment as \cite{chen2019information}. The length $T$ of the observation time window is set as 3 years, 5 years, and 7 years.

For all datasets, we randomly sample 70\% of all cascades to generate training data, 10\% for validation, and the rest for testing.

\subsection{Evaluation metric}
Following the works \cite{li2017deepcas,cao2017deephawkes,chen2019information}, we choose mean square log-transformed error (MSLE) as evaluation metric. It is defined as:
\begin{equation}
MSLE=\frac{1}{N}\underset{i=1}{\overset{N}{\mathop \sum }}\,{{\left( {{\log }_{2}}\Delta \tilde{R}_{i}^{T}-{{\log }_{2}}\Delta R_{i}^{T} \right)}^{2}},
\end{equation}
where $N$ is the total number of cascades, $\Delta R_{i}^{T}$ is the true growth size and $\Delta \tilde{R}_{i}^{T}$ is the predicted growth size.

\subsection{Baselines}
\textbf{Feature-Linear} employs linear regression to make predictions. The structural features include the average in-degree and out-degree of the cascade graph, the number of nodes, the number of leaf nodes, the number of edges. The average retweet time is used as the temporal feature.

\textbf{Feature-Deep} introduces a fully connected network to predict the cascade size from the given features.

\textbf{DeepCas} is the first deep learning architecture for information popularity prediction. It utilizes random walks to generate node sequences and uses GRU and attention mechanisms for prediction.

\textbf{DeepHawkes} combines both deep learning and point process for cascade prediction. It aims to bridge the gap between prediction and understanding of information cascades.

\textbf{CasCN} is a graph neural network based model. It samples the cascade graph as a sequence of subcascade graphs. The structural and temporal features are captured by graph convolutions and LSTM respectively.

\textbf{CasGCN} merges the activation time of the nodes with the node embedding obtained through convolutional layers. In this way, the structural and temporal characteristics are combined to make the prediction. Unfortunately, the code of CasGCN is not found publicly available. We try our best to follow the scheme of \cite{xu2020casgcn} and implement an approximate model that uses a GCN layer to learn the embedding and apply the attention mechanism to aggregate node representations.

\subsection{Parameter Settings}
For baselines, we set the dimension of user embeddings as 50, and all of the other hyper-parameters of each model are set to their default values.

The detailed setup of our model is as follows. The learning rate is initialized as 0.005. Both GCN and GAT have 2 layers. The dimension of paper representation is 32, user features include the number of bidirectional followers, followers, and friends, city number, province number, gender number, statuses, and registration time. The dimension of positional encoding ${{d}_{p}}$ is 16. 

\begin{table}[htbp]
	\centering
	\renewcommand\arraystretch{0.9}
	\caption{overall prediction performance.}
	\footnotesize
	\setlength{\tabcolsep}{0.8mm}{
		\begin{tabular}{c|ccc|ccc}
			\toprule
			\multicolumn{1}{c|}{DataSet } & \multicolumn{3}{c|}{Weibo} & \multicolumn{3}{c}{DBLP}  \\
			\midrule
			\diagbox{Models}{${T}$} & \multicolumn{1}{c|}{1 hour} & \multicolumn{1}{c|}{2 hours} & 3 hours & \multicolumn{1}{c|}{3 years} & \multicolumn{1}{c|}{5 years} & 7 years \\
			\midrule
			Feature\_Linear & 5.958 & 5.583 & 4.930 & 4.602 & 3.843 & 2.711 \\
			Feature\_Deep & 6.071 & 5.750 & 5.121 & 4.481 & 3.370 & 2.031 \\
			DeepCas & 4.790 & 4.456 & 3.761 & 3.082 & 2.233 & 1.573 \\
			DeepHawkes & 4.858 & 4.644 & 3.856 & 3.256 & 2.593 & 1.650 \\
			CasCN & 4.537 & 4.291 & 3.673 & 2.062 & 1.844 & 1.207 \\
			CasGCN & 5.268 & 4.811 & 4.076 & 1.896 & 1.589 & 1.296 \\
			CCasGNN & \textbf{4.413} & \textbf{4.201} & \textbf{3.617} & \textbf{1.792} & \textbf{1.553} & \textbf{1.146} \\
			\bottomrule
		\end{tabular}%
	}
	\label{tab:addlabel}%
\end{table}%

\subsection{Performance Comparison}
The performance comparison is shown in Table II. CCasGNN consistently outperforms the other competitive baselines for both datasets.

Compared with feature based methods, the deep learning based approaches show excellent performance, which demonstrates a clear advantage of deep learning in the popularity prediction. It is interesting to note that Feature-Linear is always better than Feature-Deep in the Weibo dataset. This suggests that increasing the number of network layers does not necessarily lead to better performance. In DeepCas, the node representation is learned by DeepWalk \cite{perozzi2014deepwalk}, and not supervised by the prediction task, which limits its performance. On the contrary, DeepHawkes learns node representation by a supervised framework. While it successfully combines the deep learning and point process, its perform is still limited due to the fact that the structural features are omitted in this model. Although the performance gap between CasCN and CCasGNN is quite small in the Weibo dataset, CCasGNN owns a higher execution efficiency as it does not use the recurrent neural network. Under the same experimental conditions, CasCN takes about 5 times as long as CCasGNN does. Meanwhile, CasCN creates a transition probability matrix when a node joins the information propagation process, so the space complexity of CasCN is $O\left( {{N}^{3}} \right)$. Our model only needs to create two adjacency matrices that feed into GAT and GCN components, whose space complexity is $O\left( {{N}^{2}} \right)$. The performance of CasGCN is unsatisfactory, since the temporal feature is only one dimension within the embeddings, and there is no layer that specifically processes it, the sequential characteristic is not fully exploited in this model.

\subsection{Ablation Study}
To better investigate the contribution of each component in CCasGNN, we implement the following variants:

\textbf{CCasGNN-GAT}: We only use the GAT for user embedding to prove the validity of collaborative prediction.

\textbf{CCasGNN-GCN}: In contrast to CCasGNN-GAT, we only use the GCN layers to learn the user embedding.

\textbf{CCasGNN-noPE}: To demonstrate the effectiveness of positional encodings, we remove them from the model.

\textbf{CCasGNN-GRU}: To explore the contribution of positional encodings to the model compared with the RNN, we add GRU and remove the positional encodings.

\begin{table}[htbp]
	\centering
	\renewcommand\arraystretch{0.9}
	\caption{prediction performance of variants.}
	\footnotesize
	\setlength{\tabcolsep}{0.7mm}{
		\begin{tabular}{c|ccc|ccc}
			\toprule
			\multicolumn{1}{c|}{DataSet } & \multicolumn{3}{c|}{Weibo} & \multicolumn{3}{c}{DBLP}  \\
			\midrule
			\diagbox{Models}{${T}$} & \multicolumn{1}{c|}{1 hour} & \multicolumn{1}{c|}{2 hours} & 3 hours & \multicolumn{1}{c|}{3 years} & \multicolumn{1}{c|}{5 years} & 7 years \\
			\midrule
			CCasGNN-GAT & 4.569 & 4.315 & 3.837 & 1.959 & 1.716 & 1.237 \\
			CCasGNN-GCN & 4.663 & 4.382 & 3.806 & 1.903 & 1.746 & 1.212 \\
			CCasGNN-noPE & 4.529 & 4.330 & 3.744 & 1.858 & 1.793 & 1.262 \\
			CCasGNN-GRU & 4.474 & 4.253 & 3.643 & 2.239 & 1.664 & 1.174 \\
			CCasGNN & \textbf{4.413} & \textbf{4.201} & \textbf{3.617} & \textbf{1.792} & \textbf{1.553} & \textbf{1.146} \\
			\bottomrule
		\end{tabular}%
	}
	\label{tab:addlabel}%
\end{table}%

Table III outlines the performance comparison among CCasGNN and its variants.

CCasGNN-GAT and CCasGNN-GCN do not show comparable performance, demonstrating the effectiveness of co-prediction of GCN and GAT. The performance of CCasGNN-noPE decreases slightly, which tells the contribution of positional encodings in our model. The performance of CCasGNN-GRU is not as good as that of CCasGNN, which can be attributed to the positional encodings of our method that involves the sequence information.

In summary, collaborative prediction and positional encodings are critical components in CCasGNN, both of which are essential in the performance improvement.

\subsection{Parameter Analysis}
We now turn to explore the impact of tunning hyper-parameter on the performance.

\begin{figure}[h]
	\centering
	\includegraphics[width=\linewidth]{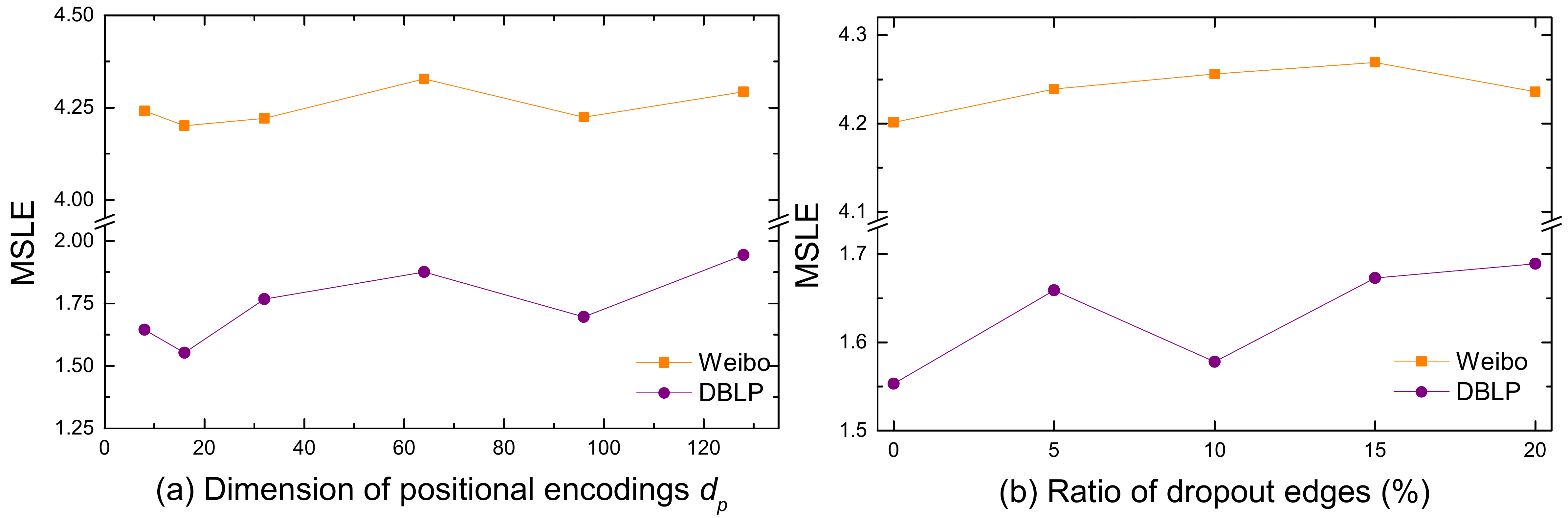}
	\caption{Impact of parameter on performance.}
\end{figure}

For the dimension of the positional encoding ${{d}_{p}}$, we set ${{d}_{p}}$ = 8, 16, 32, 64, 96, 128, and implement them on $T$ = 2 hours of Weibo dataset and $T$ = 5 years of DBLP dataset respectively, the prediction results are displayed in Fig. 2(a). We can find that the performance achieves the best when ${{d}_{p}}$ = 16. The reason may be too few dimensions lead to under-fitting, on the contrary, redundant dimensions lead to over-fitting. Another question is, to what extent the model is affected by the data quality. To answer this question, we randomly drop out a certain percentage of edges in the cascades, the dropout rate from 5\%, 10\%, 15\%, to 20\%, and predict the cascade when $T$ = 2 hours of Weibo dataset and $T$ = 5 years of DBLP dataset. Fig. 2(b) shows that the performance of our model will be slightly degraded and fluctuates in a small range, but it still shows good performance.

\section{Conclusion}
To summarize, we propose a novel collaborative cascade prediction framework based on graph neural networks, which uses collaborative GAT and GCN that takes profiles, network structure, and positional encodings into account to learn the representation of each cascade. With the cooperation of GAT and GCN, CCasGNN overcomes the individual limitations of the spectral and spatial methods. Stacking positional encodings into the layers of graph neural networks ensures that positional information is considered in every graph neural network layer and further improves the prediction performance.

\section*{Acknowledgment}
This work is supported by the National Natural Science Foundation of China (NSFC) (No. 62006198), and Industry-University-Research Innovation Fund for Chinese Universities (No. 2021ALA03016).

\bibliographystyle{IEEEtran}
\bibliographystyle{unsrt}
\bibliography{wys}

\end{document}